\documentclass[10pt, conference]{IEEEtran}
\usepackage{graphicx}
\usepackage{epstopdf}
\usepackage[nolist]{acronym}
\usepackage[normalem]{ulem}
\usepackage{romannum}
\usepackage{caption,setspace}
\usepackage{color,soul}
\usepackage{soul}
\usepackage{bbm}
\usepackage[tight,footnotesize]{subfigure}
\usepackage{mathtools,amssymb,lipsum}

\usepackage{cuted}
\usepackage{lipsum}
\usepackage{mathtools}
\usepackage{cuted}
\usepackage{amsthm}
\usepackage[table]{xcolor}
\usepackage{tabularx}
\usepackage{tabularx, ragged2e}
\usepackage[inline]{enumitem}
\usepackage[utf8]{inputenc}
\usepackage{hyperref}
\setlist{nolistsep}
\definecolor{green}{HTML}{66FF66}
\definecolor{myGreen}{HTML}{009900}
\usepackage{enumerate}
\usepackage{cleveref}
\usepackage{flushend}
\usepackage{orcidlink}
\usepackage{booktabs}    
\usepackage{xcolor}
\usepackage{amsthm}
\usepackage{enumitem}
\usepackage[utf8]{inputenc}
\usepackage{comment}
\setlist{nolistsep}

\begin{acronym} 
\acro{3GPP}{Third Generation Partnership Project}
\acro{5G}{Fifth-Generation}
\acro{AI}{Artificial Intelligence}
\acro{AIF}{Artificial Intelligence Function}
\acro{AMF}{Access and Mobility Management Function}
\acro{AUC}{Area Under The Curve}
\acro{COTS}{Commercial off-the-Shelf}
\acro{CP}{Control Plane}
\acro{CCP}{Connect Compute Platform}
\acro{DN}{Data Network}
\acro{DNN}{Data Network Name}
\acro{ETSI}{European Telecommunications Standards Institute}
\acro{GEO}{Geostationary Orbit}
\acro{GUI}{Graphical User Interface}
\acro{IFEC}{In-Flight Entertainment and Connectivity}
\acro{LEO}{Low Earth Orbit}
\acro{LiFi}{Light Fidelity}
\acro{LDN}{Local Data Network}
\acro{MEC}{Multi-access Edge Computing}
\acro{ML}{Machine Learning}
\acro{MNO}{Mobile Network Operator}
\acro{NTN}{Non-Terrestrial Network}
\acro{NR}{New Radio}
\acro{NSAP}{Network and Service Automation Platform}
\acro{NFVI}{Network Function Virtualization Infrastructure}
\acro{PDU}{Protocol Data Unit}
\acro{PLMN}{Public Land Mobile Network}
\acro{RAN}{Radio Access Network}
\acro{RDU}{Removable Display Unit}
\acro{RTT}{Round-Trip Delay}
\acro{ROC}{Receiver Operating Characteristic}
\acro{SMF}{Session Management Function}
\acro{UE}{User Equipment}
\acro{UP}{User Plane}
\acro{UPF}{User Plane Function}
\acro{SCU}{System Control Unit}
\acro{SA}{Stand Alone}
\acro{SNO}{Satellite Network Operator}
\acro{VIM}{Virtualized Infrastructure Manager}
\acro{UE}{User Equipment}
\acro{VPN}{Virtual private network}
\acro{HO}{Handover}
\acro{LEO}{Low Earth Orbit}
\acro{RAVE}{Reliable Affordable and Very Easy}
\acro{CNR}{Carrier to Noise ratio}
\acro{MSE}{Mean Squared Error}
\acro{MAE}{Mean Absolute Error}
\acro{MEO}{Medium Earth Orbit}
\acro{GPS}{Global Positioning System}
\acro{IFC}{In-Flight Connectivity}
\acro{GNSS}{Global Navigation Satellite Systems}
\acro{2D-GA}{2D Genetic Algorithm}
\end{acronym}

\begin{document}

\title{Satellite Connectivity Prediction for Fast-Moving Platforms}

\author{
		\IEEEauthorblockN
  {
  Chao Yan\IEEEauthorrefmark{1}, 
  Babak Mafakheri\IEEEauthorrefmark{1}
  }

  \IEEEauthorblockA{\IEEEauthorrefmark{1} Safran Passenger Innovations GmbH, Munich area, Germany;\\
  emails: \{name.lastname\}@zii.aero }\\
			}
\maketitle
\begin{abstract}

Satellite connectivity is gaining increased attention as the demand for seamless internet access, especially in transportation and remote areas, continues to grow. For fast-moving objects such as aircraft, vehicles, or trains, satellite connectivity is critical due to their mobility and frequent presence in areas without terrestrial coverage. Maintaining reliable connectivity in these cases requires frequent switching between satellite beams, constellations, or orbits. To enhance user experience and address challenges like long switching times, \ac{ML} algorithms can analyze historical connectivity data and predict network quality at specific locations. This allows for proactive measures, such as network switching before connectivity issues arise. In this paper, we analyze a real dataset of communication between a \ac{GEO} satellite and aircraft over multiple flights, using ML to predict signal quality. Our prediction model achieved an F1 score of 0.97 on the test data, demonstrating the accuracy of machine learning in predicting signal quality during flight. By enabling seamless broadband service, including roaming between different satellite constellations and providers,
our model addresses the need for real-time predictions of signal quality. This approach can further be adapted to automate satellite and beam-switching mechanisms to improve overall communication efficiency. The model can also be retrained and applied to any moving object with satellite connectivity, using customized datasets, including connected vehicles and trains.

\end{abstract}

\acresetall

\begin{IEEEkeywords}
Machine Learning, LEO, GEO, Handover
\end{IEEEkeywords}

\section{Introduction}

A communication satellite is an artificial satellite designed to provide communication service by transmitting signals through a transponder, creating a link between a transmitter and a receiver located at different points on Earth. These satellites play a critical role in enabling a wide range of services. 
Satellites move around the Earth due to the gravitational pull along specific paths called orbits. These orbits can vary in altitude and some other factors, depending on the satellite's purpose.  

There are various satellite providers nowadays on the market, such as ViaSat, Starlink, SES Sirius, and so on. 
Starlink satellites mainly run in \ac{LEO}, at altitudes from 340 km to 1,200 km above the Earth.  
Viasat satellites operate in GEO.  In this orbit, the satellites match the Earth's rotation and are positioned at around 35,786 km above the equator, in a fixed position to the Earth. SES Sirius is \ac{MEO} satellite provider, having their satellites operating at an altitude of around 8,000 km. 
GEO, MEO, and LEO satellites are the three commonly used communication satellite systems. 

Satellite communication is widely used in various scenarios, ranging from everyday individual use to business purposes and critical systems for overall infrastructure. The services include telecommunications, global navigation systems, autonomous systems, maritime and aviation, and other fields. Satellites' communication can widely be used for mobile phones, TV, and internet \cite{p[c]}\cite{p[d]}. Global navigation systems are utilized for providing global positioning services via \ac{GPS}\cite{p[e]}\cite{p[f]}. Autonomous systems are critical for autonomous vehicles which require high-precision GPS so that tasks such as optimizing routes, avoiding collisions, and recognizing lanes can be performed well \cite{p[a]}\cite{p[b]}. Aircraft and ships rely highly on satellites for communication, navigation and so on. Satellite communication provides internet access to crew and passengers during flights at high altitudes targeting on-board internet access, also called \ac{IFC} \cite{mafakheri2023edge,papa2023enabling}, and cruises in the middle of the ocean \cite{p[g]}\cite{p[h]}\cite{p[i]}. 
On commercial aircraft, good IFC is extremely demanding, since it can enhance passenger satisfaction to some extent which is what the airlines are pursuing \cite{p[k], SAA}.
\ac{HO} is another important term in satellite communication which means the whole step of switching an ongoing communication session from one ground station or satellite to another, during which seamless and uninterrupted connectivity should be secured. This is especially critical in aviation since the aircraft is moving at a high speed.

We have already demonstrated the usage of a LEO satellite for \ac{IFC} in \cite{mafakheri2023edge} and introduced an AI-based \ac{IFEC} system in \cite{mafakheri2023ai}. Moreover, there are several works discussing the LEO, GEO satellite networks, IFC, handover (HO), and application of ML in this domain \cite{p[l], p[m], papa2023enabling}, etc.  
Authors in \cite{p[j]} proposed a new method for IFC link allocation, which could make dynamic switches between terrestrial cellular and LEO satellite networks in real-time. 
The paper \cite{p[l]} summarizes the current ML methods for \ac{GNSS} based positioning, their advantages, disadvantages, and potential challenges. Min J \cite{p[m]} proposes a nonorthogonal multiple access (NOMA) in LEO satellite communication systems using ML techniques. The experiment results prove that the performance is comparable to the optimal one by using standard calculation. On the other hand, \cite{p[n]} proposes solutions based on ML for optimizing handover decisions in non-terrestrial networks (NTN), focusing on decreasing the issue of signaling storms during handovers. The results reveal that achieving optimal HO performance requires considering the distance between the cell center and the user, which is a critical variable in NTN.
M Chen \cite{ChenM} proposes a Q-learning-based HO scheme for LEO satellite networks that utilizes both remaining service time and signal quality to optimize the performance of handovers. Experimental results show that this method increases overall signal quality and decreases the number of HOs compared to traditional methods.
Feng L\cite{FengL}  presents a scheme by leveraging the timely and accurate orbit and position data of the aircraft to facilitate seamless HO. This scheme not only decreases the packet drop rate during HO but also minimizes data transmission delays.
Moreover, S. Mondal \cite{SabyasachiM} proposes a prediction-based solution for HO selection. They derive the cost function and constraints based on dual connectivity variables over the prediction horizon and solve the problem using a \ac{2D-GA} to achieve the best HO solution. The results indicate that network densification combined with the predictive control model improves overall aircraft performance.

Although existing studies propose methods to improve satellite handover, a prediction model specifically tailored to moving platforms can significantly enhance satellite communication handover mechanisms. To the best of our knowledge, no prior research has analyzed historical data to develop a prediction model for satellite connectivity of moving platforms.
In this context, our objective is to create a satellite-aircraft link quality prediction model that provides comprehensive analysis into link conditions throughout an entire flight across all potential geographic coordinates. The primary goal of this model is to predict the satellite signal quality in different geographical coordinates which helps enable an automated handover mechanism capable of making network switching decisions based on the aircraft’s location and other real-time data such as weather conditions during flight.

The main contribution of this paper is the development of an accurate prediction model capable of forecasting the signal quality of GEO satellites on moving aircraft. This model enables airline operators to take proactive measures before encountering issues related to connection failures, such as beam-switching or network switching. By utilizing the most relevant features for optimal signal quality prediction, the model is categorized into high-altitude and low-altitude scenarios, with low-altitude scenarios incorporating weather data as an influential factor. This adaptability allows the model to be applied to other moving objects on the ground, such as cars and trains. 

The remainder of the article is structured as follows: Section II discusses the methodology, including dataset processing, metrics selection, and training. Section III presents the experimental results, and Section IV concludes the work with a discussion of future directions.

\section{Methodology}
This section outlines the methods used to analyze the dataset and develop models for predicting the received signal quality of a \ac{GEO} satellite on an aircraft during flight. We detail the steps involved in data processing, the criteria for selecting performance metrics, and the machine learning experiments conducted. The results of the prediction model are also discussed with a focus on the model's effectiveness in predicting signal quality.

\subsection{Dataset Processing}

The dataset used in this study consists of real-world data collected from several flights, capturing detailed logs of aircraft operations. It contains a comprehensive set of parameters, including the status of the aircraft’s software and hardware as well as all relevant communication metrics throughout each flight. The logging process begins after the aircraft reaches a certain altitude post-takeoff and records data at one-minute intervals. Thus, it provides a continuous snapshot of communication conditions for nearly the entire duration of each flight.

Within this dataset, we selected a subset for training and experimentation, comprising seven months of real-world data from multiple flights conducted in 2023. The data was extracted from a commercial database logged within the \ac{RAVE} \cite{RAVE} system. The initial dataset contains over 10 million entries, with each log containing up to 81 features. These features include numeric values, such as the aircraft's altitude, longitude, and latitude, as well as categorical features, such as TailNumber, AirlineCode, DepartureAirport, and ArrivalAirport. Time-related features, like LogDate, FlightStartTime, and FlightEndTime, are stored in time format. Of particular interest is the downlink carrier feature, \ac{CNR}, measured in dB, which serves as the target feature for our machine learning model.

\begin{figure}[ht!]
	\centering
    \includegraphics[width=\linewidth]{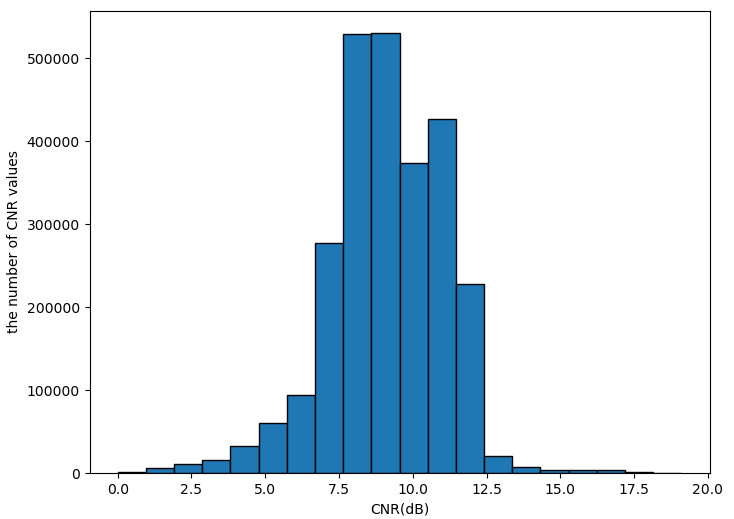}
	\caption{Distribution of CNR Values}
	\label{fig:CNR}
\end{figure}

First, we analyzed the distribution of \ac{CNR} values. As shown in \Cref{fig:CNR}, most CNR values range between 8 and 10 dB, with a mean of 8.82 dB and a median of 8.8 dB.
Since many features remained constant throughout the flight, we filtered out these static features and selected 12 key features for our experiment, including the target feature. To further structure our analysis, we created two distinct datasets. Dataset 1 focuses on the top five airline routes with the highest number of recorded logs, primarily consisting of long-haul flights. Dataset 2 contains the full dataset, covering all recorded flights across various routes. The purpose of using two datasets is to compare the models’ performances between specific flight paths (Dataset 1) and general flight paths (Dataset 2).


\begin{figure*}[htp!]
	\centering
        \includegraphics[width=\textwidth]{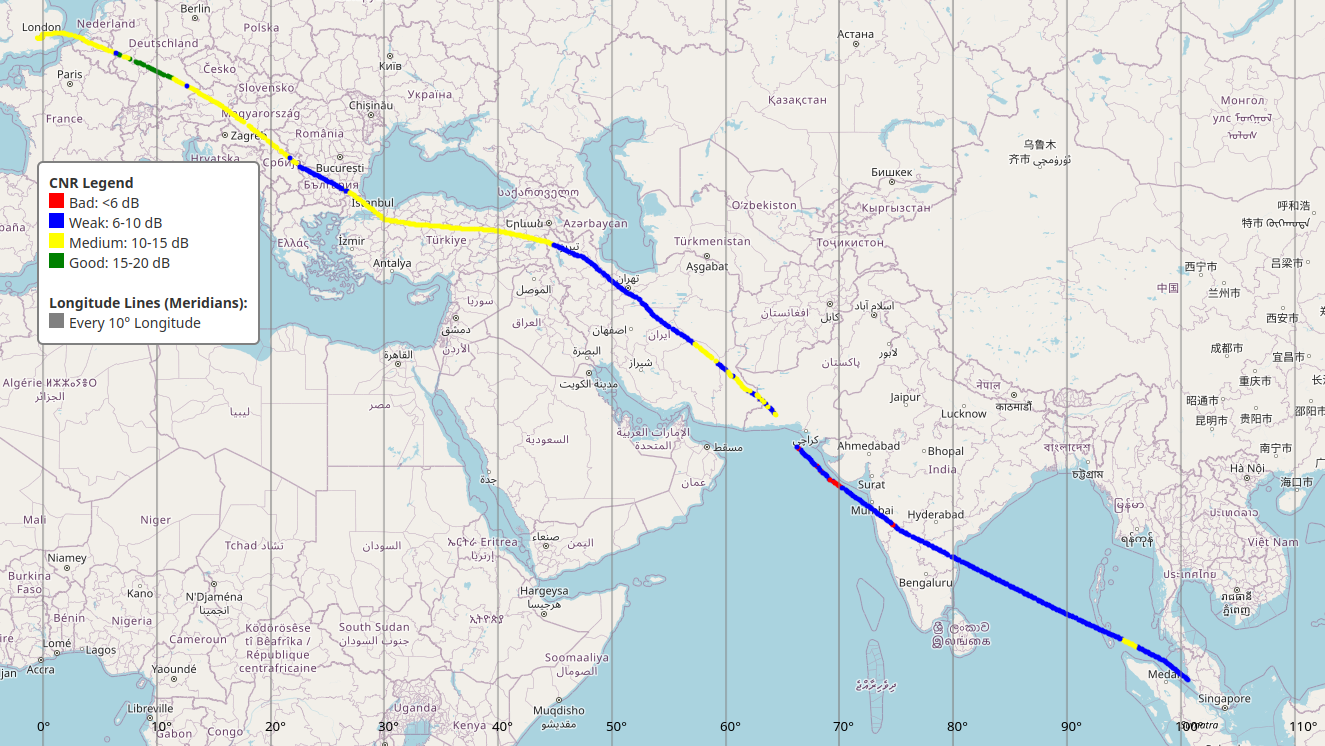}
	\caption {CNR values of some flight from Singapore to London}
	\label{fig:SIN_Lon}
\end{figure*}

\subsection{ML Experiment Conduction \& Metrics Selection}
We initially conducted the experiment using the original numeric CNR values, building regression models to predict these values, with \ac{MSE} and \ac{MAE} as our validation metrics. MSE and MAE are commonly used in regression tasks. However, the results for both MAE and MSE were around 0.2, which is high and not ideal. Our ultimate goal is to predict CNR values to trigger a network switching, including HO when necessary. This means that as long as the CNR reaches a specific threshold or range, a switch should be performed. In other words, precise \ac{CNR} values are not required directly for this purpose; instead, we can predict the range of CNR values to determine when a switch should be activated and this could increase the error-tolerant rate in our use case. Therefore, we transformed the continuous CNR values into categorical ranges, where predicting the correct category would be sufficient for identifying when to activate HO.  
Dividing the CNR values into categories presented another challenge. Different modulation types and coding rates have different CNR requirements. In our case, we convert the CNR values into 4 categories: Good 15-20 dB; Medium 10-15 dB; Weak 6-10 dB; and Bad \textless 6 dB, which is based on our use case.
Our use case then turns into a multi-class classification problem with imbalanced dataset labels. In such scenarios, accuracy or precision alone may not provide a comprehensive assessment of model performance, as they do not account for the imbalance in class distribution. Instead, the F1 score is more appropriate, as it balances precision and recall, offering a better evaluation of the model's effectiveness across all classes.

We then randomly selected a sample flight and plotted the CNR values along the flight path. Fig.\ref{fig:SIN_Lon} shows the CNR values of the selected flight from Singapore to London, with data recorded every minute. We noticed that the CNR values fluctuated during the flight, with certain segments showing low performance, falling into the Medium or Bad categories, or even missing entirely. Several factors could contribute to these fluctuations, including weather conditions at low altitudes, restricted areas, or other contributing factors.  

In our experiment results, we observed that the prediction model's performance was worse at low altitudes compared to higher altitudes. We believe the lack of weather data in our dataset could be a vital issue leading to this. To address and verify the impact of weather data on model performance, we divided our model into two sub-models: one for low-altitude data, which incorporates weather information, and another for high-altitude data, which excludes weather data. We sought external sources and identified some open-source weather datasets for our second sub-model. Historical weather records are typically available on an hourly basis, whereas our dataset is recorded at one-minute intervals. After evaluating options, we selected a historical weather data provider \cite{weather1}.
To integrate the weather data with our primary dataset, we utilized geo-coordinates and timestamps as common reference points. For each instance in our dataset, we identified corresponding weather conditions by matching its recorded coordinates and timestamp to the nearest available weather data point (precision to about 10 km). This approach ensured alignment between the two datasets and allowed us to incorporate environmental factors into our analysis. For model development, we utilized the open-source H2O AutoML framework\cite{h2o} in our local environment. H2O AutoML simplifies the machine learning model-building process and provides a user-friendly interface that enables data scientists to construct reliable models efficiently. It includes built-in algorithms such as H2O Gradient Boosting Machines, XGBoost Gradient Boosting Machines, Generalized Linear Models, Distributed Random Forests, Deep Learning, and Stacked Ensemble models (generated by combining other base models). 

\section{Experiments Result}



We conducted multiple experiments using two datasets under varied conditions, as shown in Table I. This table presents datasets filtered by different altitude thresholds, unique satellite identifiers, and the presence or absence of weather information.

First, Indexes 1 to 4 compare various altitude thresholds within Dataset 1. We observed that the prediction model becomes more reliable at higher altitudes, achieving its best result with an F1 score of 0.97349 when filtering for altitudes above 6000 m. Given these findings, we assumed that incorporating weather data might improve prediction accuracy, especially at lower altitudes where performance declines.

To test this, we used Index pairs 5 and 6 for  Dataset 1, as well as 7 and 8 for  Dataset 2, to examine the impact of weather information on predictions at lower altitudes. As expected, we saw a slight increase in F1 scores for both datasets when weather data was included, indicating that weather conditions positively influence prediction accuracy in these scenarios.

Meanwhile, the F1 score of the model using Dataset 1 (Index 1) is higher than that using Dataset 2 (Index 9), and they have the same altitude threshold (altitude \textgreater 6000 m). The same trend happens also in Indexes 5 and 7, with the identical altitude threshold, the F1 score of Index 5 is higher than Index 7.
This is likely because Dataset 1 contains more similar data (focusing on long-haul flights), whereas Dataset 2 is more generalized.  

Lastly, Indexes 10 to 14 correspond to models trained on data from individual satellites. Among these, CNR predictions based on Dataset 2 from Satellite I5F3 outperform those from the other satellites. However, further research is required to determine the underlying reasons for this performance.

\begin{table}[h]
    \centering
    \caption{F1 Scores of Various ML Models}
    \resizebox{0.5\textwidth}{!}{ 
        \begin{tabular}{@{}|l|c|c|c|@{}} 
            \toprule
            Index & Dataset & Dataset Size & F1  \\ \midrule
            1&Dataset 1 with altitude \textgreater 6000m  & 811,233 & 0.97349 \\ \midrule
            2&Dataset 1 with altitude \textgreater 5000m  & 817,817 & 0.97065 \\ \midrule
            3&Dataset 1 with altitude \textgreater 4000m  & 824,659 & 0.97172 \\ \midrule
            4&Dataset 1 with altitude \textgreater 3000m  & 832,284 & 0.97133 \\ \midrule
            5&Dataset 1 with altitude \textless 3000m & 161,388& 0.94569 \\ \midrule
            6&Dataset 1 with altitude \textless 3000m + 4 weather data & 161,388 & 0.96054 \\ \midrule
            7&Dataset 2 with altitude \textless 3000m & 230,685 & 0.90216 \\ \midrule
            8&Dataset 2 with altitude \textless 3000m + 4 weather data & 230,685 & 0.91685 \\ \midrule
            9&Dataset 2 with altitude \textgreater 6000m & 2,617,623 & 0.96802 \\ \midrule
            10&Dataset 2 with altitude \textgreater 6000m for I5F1 & 804,578 & 0.95916 \\ \midrule
            11&Dataset 2 with altitude \textgreater 6000m for I5F2 & 563,815 & 0.95421 \\ \midrule
            12&Dataset 2 with altitude \textgreater 6000m for I5F3 & 434,963 & 0.96804 \\ \midrule
            13&Dataset 2 with altitude \textgreater 6000m for I5F4 & 226,659 & 0.95700 \\ \midrule
            14&Dataset 2 with altitude \textgreater 6000m for GX5 & 37,353 & 0.914340 \\

            \bottomrule
        \end{tabular}
    }
    \label{tab:t3}
\end{table}

\section{Conclusion}

In this paper, we developed and evaluated machine learning models to predict CNR values during flights, with the ultimate goal of improving satellite handover mechanisms. Our results demonstrate that at higher altitudes, the model's performance is more reliable, achieving an F1 score of 0.97349 with selected flights. This indicates the model's high effectiveness in predicting signal quality, which could be potentially used in customized flight paths. Furthermore, we found that incorporating weather data at lower altitudes slightly improved the F1 score, confirming that external environmental factors like weather play a role in signal quality fluctuations. 

By comparing models trained on different datasets, we observed that Dataset 1, which contains more homogenous data from long-haul flights, outperforms the more generalized Dataset 2, particularly when using the same altitude thresholds. Additionally, models trained with data from a single satellite revealed that predictions using Satellite I5F3 data outperformed those from other satellites, indicating that some satellites might provide a more stable link quality for aircraft. 

Overall, this work proves the viability of using machine learning for predicting link quality and highlights the potential for these models to support the automation of satellite handover processes. This approach could significantly reduce signal disruptions, improve in-flight connectivity reliability, and enhance passenger experience and operational efficiency by predicting signal quality in advance and securing the best link. Future work will focus on automating handover decisions based on these predictions, integrating both satellite and weather data for real-time optimization. 

\section*{Acknowledgment}
This work received support from the German Federal Ministry for Economic Affairs and Energy (BMWi) through the German Aeronautical Research Program LuFo - the INTACT project (grant agreement No. 20D2128F). 
 
\bibliographystyle{IEEEtran}
\bibliography{main}

\end{document}